\documentclass[10pt,twocolumn,letterpaper]{article}

\usepackage[utf8]{inputenc}
\usepackage[T1]{fontenc}
\usepackage{iccv}
\usepackage{times}
\usepackage{textcomp}
\usepackage{epsfig}
\usepackage{graphicx}
\usepackage{amsmath}
\usepackage{amssymb}
\usepackage[numbers,sort]{natbib}
\usepackage{float}
\usepackage{booktabs}
\usepackage{capt-of}
\usepackage[toc,page]{appendix}
\usepackage[moderate,mathdisplays=normal,wordspacing=normal]{savetrees}
\makeatletter
\renewcommand\paragraph{\@startsection{paragraph}{4}{\z@}%
	{0.75ex \@plus.5ex \@minus.2ex}
	{-1em}%
	{\normalfont\normalsize\bfseries}}
\makeatother

\usepackage[pagebackref=true,breaklinks=true,colorlinks,bookmarks=false]{hyperref}
\iccvfinalcopy 

\newcommand{\abs}[1]{\left\lvert#1\right\rvert}
\newcommand{\norm}[1]{\left\lVert#1\right\rVert}

\AtBeginDocument{%
\setlength\abovedisplayskip{5pt plus 1pt minus 2.5pt} 
\setlength\belowdisplayskip{5pt plus 1pt minus 2.5pt} 
\setlength\abovedisplayshortskip{0pt plus 3pt} 
\setlength\belowdisplayshortskip{3pt plus 1.5pt minus 1.5pt} 
}

\ificcvfinal\fi
\begin{document}
\title{HoloGAN: Unsupervised Learning of 3D Representations From Natural Images}

\author{%
	Thu Nguyen-Phuoc $^\text{1}$\quad%
	Chuan Li $^\text{2}$\quad%
	Lucas Theis $^\text{3}$\quad%
	Christian Richardt $^\text{1}$\quad%
	Yong-Liang Yang $^\text{1}$\quad%
	\\[.75em]%
	$^\text{1}$ University of Bath\quad%
	$^\text{2}$ Lambda Labs\quad%
	$^\text{3}$ Twitter%
}

\twocolumn[{
	\renewcommand\twocolumn[1][]{#1}
	\maketitle
	\vspace{-0.5em}
	{\centering
		\includegraphics[width=\linewidth]{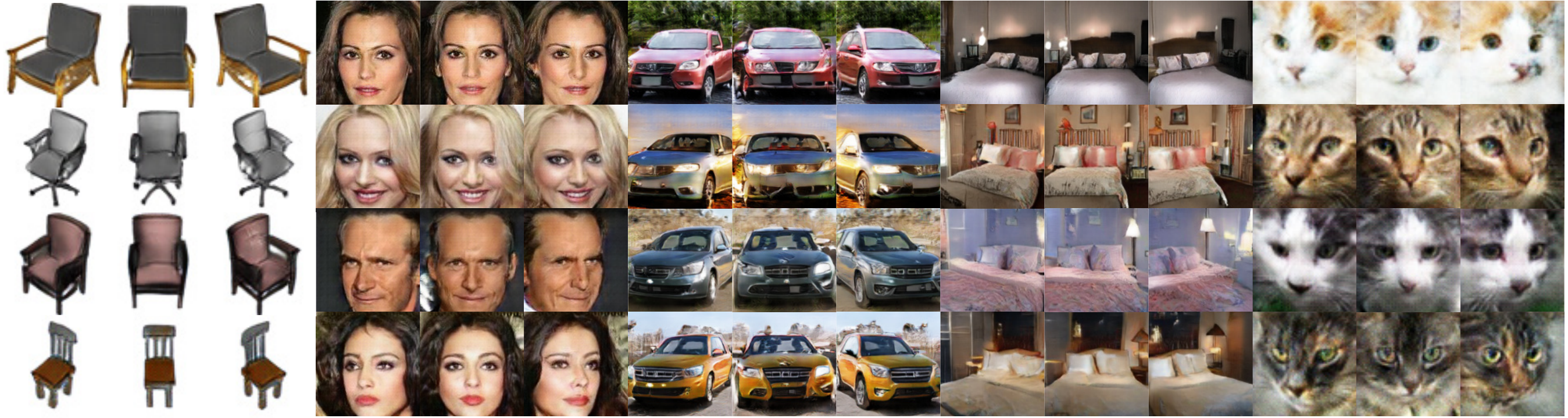}
		\captionof{figure}{%
			HoloGAN learns to separate pose from identity (shape and appearance) only from unlabelled 2D images without sacrificing the visual fidelity of the generated images.
			All results shown here are sampled from HoloGAN for the same identities in each row but in different poses.
		}
		\label{fig:moneyShot}
	}
	\vspace{0.5cm}
}]

\begin{abstract}
We propose a novel generative adversarial network (GAN) for the task of unsupervised learning of 3D representations from natural images.
Most generative models rely on 2D kernels to generate images and make few assumptions about the 3D world. 
These models therefore tend to create blurry images or artefacts in tasks that require a strong 3D understanding, such as novel-view synthesis. 
HoloGAN instead learns a 3D representation of the world, and to render this representation in a realistic manner.
Unlike other GANs, HoloGAN provides explicit control over the pose of generated objects through rigid-body transformations of the learnt 3D features. 
Our experiments show that using explicit 3D features enables HoloGAN to disentangle 3D pose and identity, which is further decomposed into shape and appearance, while still being able to generate images with similar or higher visual quality than other generative models.
HoloGAN can be trained end-to-end from unlabelled 2D images only. 
In particular, we do not require pose labels, 3D shapes, or multiple views of the same objects. 
This shows that HoloGAN is the first generative model that learns 3D representations from natural images in an entirely unsupervised manner.

\end{abstract}

\section{Introduction}
\label{sec:introduction}

Learning to understand the relationship between 3D objects and 2D images is an important topic in computer vision and computer graphics.
In computer vision, it has applications in fields such as robotics, autonomous vehicles or security.
In computer graphics, it benefits applications in both content generation and manipulation.
This ranges from photorealistic rendering of 3D scenes or sketch-based 3D modelling, to novel-view synthesis or relighting.

\begin{figure}
	\centering
	\includegraphics[width=\linewidth]{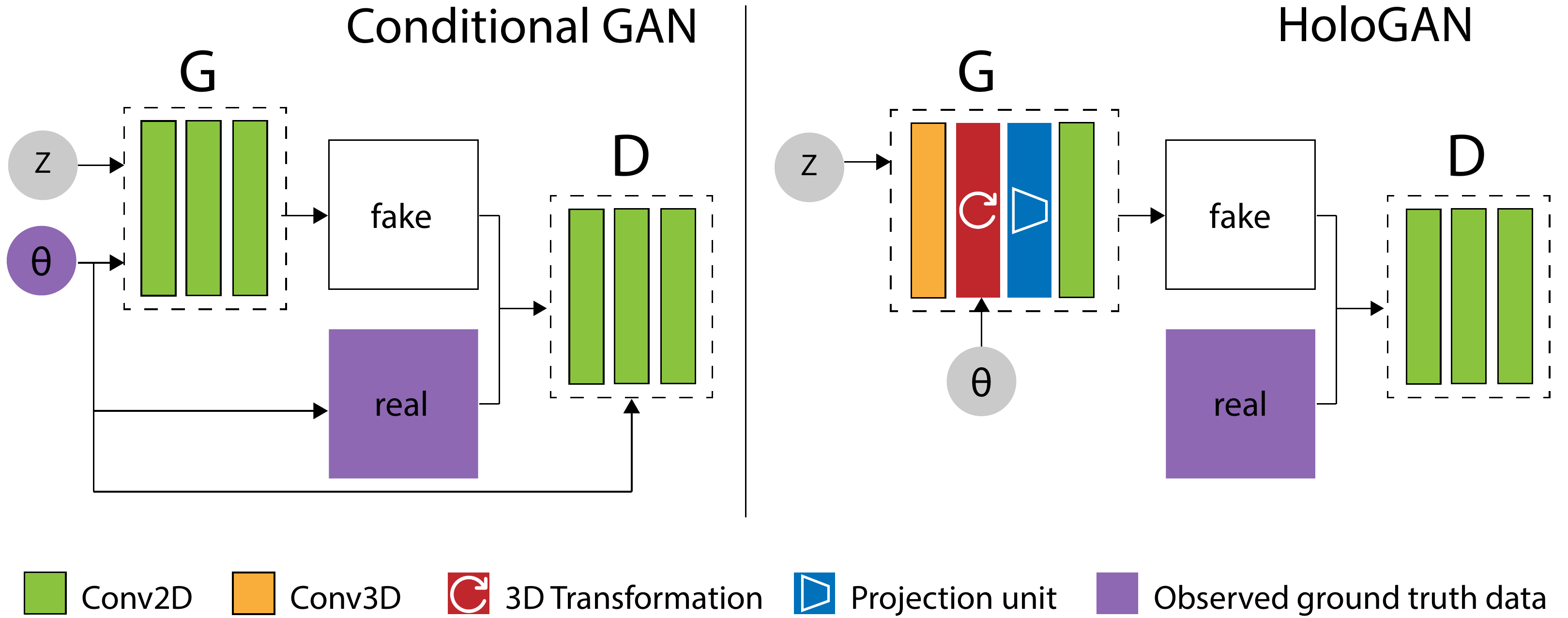}
	\caption{\label{fig:schematics}%
		Comparison of generative image models. Data given to the discriminator are coloured purple.
		\textbf{Left:} In conditional GANs, the pose is observed and the discriminator is given access to this information. \textbf{Right:} HoloGAN does not require pose labels during training and the discriminator is not given access to pose information.
	}
\end{figure}

Recent generative image models, in particular, generative adversarial networks (GANs), have achieved impressive results in generating images of high resolution and visual quality \cite{karras2018, karras2019, brock2018large, NIPS2018_7627, Han18}, while their conditional versions have achieved great progress in image-to-image translation \cite{Isola2017, sangkloy2016scribbler}, image editing \cite{Dorta2018TheGT, ding2017exprgan, Zhang_2018_ECCV} or motion transfer \cite{Chan2018dance, kim2018DeepVideo}.
However, GANs are still fairly limited in their applications, since they do not allow explicit control over attributes in the generated images, while conditional GANs need labels during training (Figure \ref{fig:schematics} left), which are not always available.

Even when endowed with labels like pose information, current generative image models still struggle in tasks that require a fundamental understanding of 3D structures, such as novel-view synthesis from a single image.
For example, using 2D kernels to perform 3D operations, such as out-of-plane rotation to generate novel views, is very difficult. 
Current methods either require a lot of labelled training data, such as multi-view images or segmentation masks \cite{tvsn_cvpr2017, sun2018multiview}, or produce blurry results \cite{Dosovitskiy2014, Kulkarni2015, Eslami1204, Yan2016, TDB16a}.
Although recent work has made efforts to address this problem by using 3D data \cite{NIPS2018_8014, NIPS2018_7297}, 3D ground-truth data are very expensive to capture and reconstruct.
Therefore, there is also a practical motivation to directly learn 3D representations from unlabelled 2D images.

Motivated by these observations, we focus on designing a novel architecture that allows unsupervised learning of 3D representations from images, enabling direct manipulation of view, shape and appearance in generative image models (Figure \ref{fig:moneyShot}).
The key insight of our network design is the combination of a strong inductive bias about the 3D world with deep generative models to learn better representations for downstream tasks.
Conventional representations in computer graphics, such as voxels and meshes, are \emph{explicit} in 3D and easy to manipulate via, for example, rigid-body transformations. 
However, they come at the cost of memory inefficiency or ambiguity in how to discretise complex objects. 
As a result, it is non-trivial to build generative models with such representations directly \cite{Riegler2017OctNet, NIPS2017_7095, COMA:ECCV2018}. 
\emph{Implicit} representations, such as high-dimensional latent vectors or deep features, are favoured by generative models for being spatially compact and semantically expressive. 
However, these features are not designed to work with explicit 3D transformations \cite{Chen2016, higgins2017, Eslami1204, Rezende2016, rajeswar2019pixscene}, which leads to visual artefacts and blurriness in tasks such as view manipulation.

We propose HoloGAN, an unsupervised generative image model that learns representations of 3D objects that are not only explicit in 3D but also semantically expressive.
Such representations can be learnt directly from unlabelled natural images.
Unlike other GAN models, HoloGAN employs both 3D and 2D features for generating images.
HoloGAN first learns a 3D representation, which is then transformed to a target pose, projected to 2D features, and rendered to generate the final images (Figure \ref{fig:schematics} right).
Different from recent work that employs hand-crafted differentiable renderers \cite{Loper2014, kato2018renderer, insafutdinov18pointclouds, NIPS2018_7297, Sitzmann2018, Li2018, henderson18}, HoloGAN learns perspective projection and rendering of 3D features from scratch using a projection unit \cite{NIPS2018_8014}.
This novel architecture enables HoloGAN to learn 3D representations directly from natural images for which there are no good hand-crafted differentiable renderers.
To generate new views of the same scene, we directly apply 3D rigid-body transformations to the learnt 3D features, and visualise the results using the neural renderer that is jointly trained. 
This has been shown to produce sharper results than performing 3D transformations in high-dimensional latent vector space \cite{NIPS2018_8014}.

HoloGAN can be trained end-to-end in an unsupervised manner using only unlabelled 2D images, without any supervision of poses, 3D shapes, multiple views of objects, or geometry priors such as symmetry and smoothness over the 3D representation that are common in this line of work \cite{BarronTPAMI2015, rajeswar2019pixscene, cmrKanazawa18}.
To the best of our knowledge, HoloGAN is the first generative model that can learn 3D representations directly from \textit{natural} images in a purely unsupervised manner. In summary, our main technical contributions are:
\begin{itemize}
	\item  A novel architecture that combines a strong inductive bias about the 3D world
	with deep generative models to learn disentangled
	representations (pose, shape, and appearance) of 3D objects from images. The representation
	is explicit in 3D and expressive in semantics.
	\item An unconditional GAN that, for the first time, allows native support for view manipulation 
	without sacrificing visual image fidelity.
	\item An unsupervised training approach that enables disentangled representation learning without using labels.
\end{itemize}
 
\section{Related work}
\label{sec:relatedwork}

\begin{figure*}
	\centering
	\includegraphics[width=\linewidth]{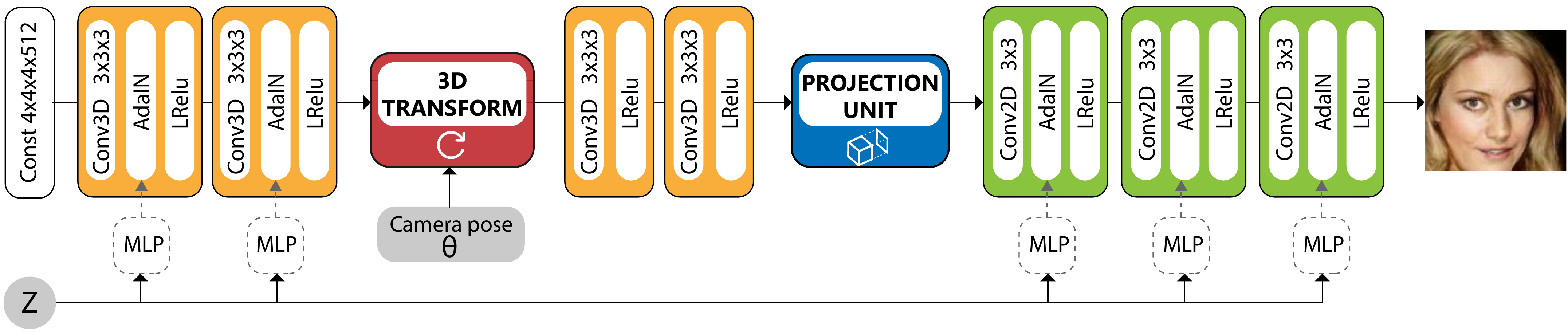}
	\caption{\label{fig:RenderGANArchitecture}%
		HoloGAN's generator network: we employ 3D convolutions, a 3D rigid-body transformation, the projection unit and 2D convolutions. We also remove the traditional input layer from $\mathbf{z}$, and start from a learnt constant 4D tensor.
		The latent vector $\textbf{z}$ is instead fed through multilayer perceptrons (MLPs) to map to the affine transformation parameters for adaptive instance normalisation (AdaIN).
		Inputs are coloured gray.
	}
\end{figure*}

HoloGAN is at the intersection of GANs, structure-aware image synthesis and disentangled representation learning.
In this section, we review related work in these areas.

\subsection{Generative adversarial networks}
GANs learn to map samples from an arbitrary latent distribution to data that fool a discriminator network into categorising them as real data \cite{NIPS2014_5423}.
Most recent work on GAN architectures has focused on improving training stability or visual fidelity of the generated images, such as multi-resolution GANs \cite{karras2018, han2017stackgan}, or self-attention generators \cite{Han18, NIPS2018_7627}.
However, there is far less work on designing GAN architectures that enable unsupervised disentangled representation learning which allows control over attributes of the generated images.
By injecting random noise and adjusting the ``style'' of the image at each convolution, StyleGAN \cite{karras2019} can separate fine-grained variation (e.g., hair, freckles) from high-level features (e.g., pose, identity), but does not provide explicit control over these elements.
A similar approach proposed by \citet{chen2018on} shows that this network design also achieves more training stability.
The success of these approaches indicates that network architecture can be more important for training stability and image fidelity than the specific choice of GAN loss.
Therefore, we also focus on architecture design for HoloGAN, but with the goal of learning to separate pose, shape and appearance, and to enable direct manipulation of these elements.

\subsection{3D-aware neural image synthesis}

Recent work in neural image synthesis and novel-view synthesis has found success in improving the fidelity of the generated images with 3D-aware networks.
Work that uses geometry templates achieves great improvements in image fidelity \cite{Kossaifi2018, GengSZWZ2018}, but does not generalise well to complex datasets that cannot be described by a template.
RenderNet \cite{NIPS2018_8014} introduces a differentiable renderer using a convolutional neural network (CNN) that learns to render 2D images directly from 3D shapes.
However, RenderNet requires 3D shapes and their corresponding rendered images during training.
Other approaches learn 3D embeddings that can be used to generate new views of the same scene without any 3D supervision \cite{Rhodin2018,Sitzmann2018}. 
However, while \citet{Sitzmann2018} require multiple views and pose information as input, \citet{Rhodin2018} require supervision from paired images, background segmentation and pose information.
Aiming to separate geometry from texture, visual object networks (VONs) \cite{NIPS2018_7297} first sample 3D objects from a 3D generative model, render these objects using a hand-crafted differentiable layer to normal, depth and silhouette maps, and finally apply a trained image-to-image translation network. However, VONs need explicit 3D data for training, and only work for single-object images with a simple white background.
Our HoloGAN also learns a 3D representation and renders it to produce a 2D image but without seeing any 3D shapes, and works with real images containing complex backgrounds and multi-object scenes.

The work closest to ours is Pix2Scene \cite{rajeswar2019pixscene}, which learns an implicit 3D scene representation from images, also in an unsupervised manner. However, this method maps the implicit representation to a surfel representation for rendering, while HoloGAN uses an explicit 3D representation with deep voxels.
Additionally, using a hand-crafted differentiable renderer, Pix2Scene can only deal with simple synthetic images (uniform material and lighting conditions).
HoloGAN, on the other hand, learns to render from scratch, and thus works for more complex natural images.

\subsection{Disentangled representation learning}
\label{disentangled}
The aim of disentangled representation learning is to learn a factorised representation, in which changes in one factor only affect the corresponding elements in the generated images, while being invariant to other factors.
Most work in disentangled learning leverages labels provided by the dataset \cite{tran2017representation, Bao_2018_CVPR, Reed2014} or benefits from set supervision (e.g., videos or multiple images of the same scene; more than two domains with the same attributes) \cite{Kulkarni2015, NIPS2017_7028, Eslami1204}.

Recent efforts in unsupervised disentangled representation learning, such as $\beta$-VAE \cite{higgins2017} or InfoGAN \cite{Chen2016, jeon2019ibgan}, have focused mostly on designing loss functions. However, these models are sensitive to the choice of priors, provide no control over what factors are learned, and do not guarantee that the learnt disentangled factors are semantically meaningful.
Moreover, $\beta$-VAE comes with a trade-off between the quality of the generated images and the level of disentanglement.
Finally, these two methods struggle with more complicated datasets (natural images with complex backgrounds and lighting).
In contrast, by redesigning the architecture of the generator network, HoloGAN learns to successfully separate pose, shape and appearance, as well as providing explicit pose control and enables shape/appearance editing even for more complex natural image datasets.

\section{Method}
\label{sec:method}
To learn 3D representations from 2D images without labels, HoloGAN extends traditional unconditional GANs by introducing a strong inductive bias about the 3D world into the generator network. Specifically, HoloGAN generates images by learning a 3D representation of the world and to render it realistically such that it fools the discriminator.
View manipulation therefore can be achieved by directly applying 3D rigid-body transformations to the learnt 3D features.
In other words, the images created by the generator are a \emph{view-dependent} mapping from a learnt 3D representation to the 2D image space.
This is different from other GANs which learn to map a noise vector $\mathbf{z}$ directly to 2D features to generate images.

Figure \ref{fig:RenderGANArchitecture} illustrates the generator architecture of HoloGAN: HoloGAN first learns a 3D representation (assumed to be in a canonical pose) using 3D convolutions (Section \ref{sec:template}), transforms this representation to a certain pose, projects and computes visibility using the projection unit (Section \ref{sec:view_transformed}), and computes shaded colour values for each pixel in the final images with 2D convolutions.
HoloGAN shares many rendering insights with RenderNet \cite{NIPS2018_8014}, but works with natural images, and needs neither pre-training of the neural renderer nor paired 3D shape–2D image training data.

During training, we sample random poses from a uniform distribution and transform the 3D features using these poses before rendering them to images.
We assume every image has a corresponding single global pose, and show that this assumption still works with images of multiple objects.
This random pose perturbation pushes the generator network to learn a disentangled representation that is suitable for both 3D transformation and generating images that can fool the discriminator.
While pose transformation could be learnt from data, 
we provide this operation, which is differentiable and straightforward to implement, explicitly to HoloGAN.
Using explicit rigid-body transformations for novel-view synthesis has been shown to produce sharper images with fewer artefacts \cite{NIPS2018_8014}.
More importantly, this provides an inductive bias towards representations that are compatible with \emph{explicit} 3D rigid-body transformations.
As a result, the learnt representations are explicit in 3D and disentangled between pose and identity.

\citet{Kulkarni2015} categorize the learnt disentangled representation into \emph{intrinsic} and \emph{extrinsic} elements. While intrinsic elements describe shape, appearance, etc., extrinsic elements describe pose (elevation, azimuth) and lighting (location, intensity).
The design of HoloGAN naturally lends itself to this separation by using more inductive biases about the 3D world: the adoption of a native 3D transformation, which controls the pose (shown as $\mathbf{\theta}$ in Figure \ref{fig:RenderGANArchitecture}) directly, to the learnt 3D features, which control the identity (shown as $\mathbf{z}$ in Figure \ref{fig:RenderGANArchitecture}).

\subsection{Learning 3D representations}
\label{sec:template}
HoloGAN generates 3D representations from a learnt constant tensor  (see Figure \ref{fig:RenderGANArchitecture}). 
The random noise vector $\mathbf{z}$ instead is treated as a “style” controller, 
and mapped to affine parameters for adaptive instance normalization (AdaIN) \cite{huang2017adain} after each convolution using a  multilayer perceptron (MLP) $f: \mathbf{z} \rightarrow \gamma(\mathbf{z}),~\sigma(\mathbf{z})$.

Given some features $\mathbf{\Phi}_l$ at layer $l$ of an image $\mathbf{x}$  and the noise ``style'' vector $\mathbf{z}$, AdaIN is defined as:
\begin{align}
\text{AdaIN}(\mathbf{\Phi}_l(\mathbf{x}), ~\mathbf{z}) = \sigma(\mathbf{z}) \left(\frac{\mathbf{\Phi}_l(\mathbf{x}) - \mu(\mathbf{\Phi}_l(\mathbf{x}))}{\sigma(\mathbf{\Phi}_l(\mathbf{x}))}\right) + \gamma(\mathbf{z}) \text{.}
\end{align}
This can be viewed as generating images by transformation of a template (the learnt constant tensor) using AdaIN to match the mean and standard deviation of the features at different levels $l$ (which are believed to describe the image ``style'') of the training images.
Empirically, we find this network architecture can disentangle pose and identity much better than those that feed the noise vector $\mathbf{z}$ directly to the first layer of the generator.

HoloGAN inherits this style-based strategy from StyleGAN \cite{karras2019} but is different in two important aspects. 
Firstly, HoloGAN learns 3D features from a learnt 4D constant tensor (size 4$\times$4$\times$4$\times$512, where the last dimension is the feature channel) before projecting them to 2D features to generate images, while StyleGAN only learns 2D features.
Secondly, HoloGAN learns a disentangled representation by combining 3D features with rigid-body transformations during training, while StyleGAN injects independent random noise into each convolution. 
StyleGAN, as a result, learns to separate 2D features into different levels of detail, depending on the feature resolution, from coarse (e.g., pose, identity) to more fine-grained details (e.g., hair, freckles).
We observe a similar separation in HoloGAN.
However, HoloGAN further separates pose (controlled by the 3D transformation), shape (controlled by 3D features), and appearance (controlled by 2D features).

It is worth highlighting that to generate images at 128$\times$128 (same as VON), we used a deep 3D representation of size up to 16$\times$16$\times$16$\times$64.
Even with such limited resolution, HoloGAN can still generate images with competitive quality and more complex backgrounds than other methods that use full 3D geometry such as VON's voxel grid of resolution 128$\times$128$\times$128$\times$1 \cite{NIPS2018_7297}.

\subsection{Learning with view-dependent mappings}
\label{sec:view_transformed}
In addition to adopting 3D convolutions to learn 3D features, 
during training, we introduce more bias about the 3D world by transforming these learnt features to random poses before projecting them to 2D images. 
This random pose transformation is crucial to guarantee that HoloGAN learns a 3D representation that is disentangled and can be rendered from all possible views,
as also observed by \citet {tran2017representation} in DR-GAN.
However, HoloGAN performs explicit 3D rigid-body transformation, while DR-GAN performs this using an implicit vector representation.

\paragraph{Rigid-body transformation}
We assume a virtual pinhole camera that is in the canonical pose (axis-aligned and placed along the negative z-axis) relative to the 3D features being rendered.
We parameterise the rigid-body transformation by 3D rotation, scaling followed by trilinear resampling.
Although translation is inherently supported by our framework, we did not use it in this work. 
Assuming the up-vector of the object coordinate system is the global y-axis, rotation comprises rotation around the y-axis (azimuth) and the x-axis (elevation). 
Details on ranges for pose sampling are included in the supplemental document.

\paragraph{Projection unit}
In order to learn meaningful 3D representations from just 2D images, HoloGAN learns a differentiable projection unit \cite{NIPS2018_8014} that reasons over occlusion.
In particular, the projection unit receives a 4D tensor (3D features), and returns a 3D tensor (2D features).
 
Since the training images are captured with different perspectives,
HoloGAN needs to learn perspective projection.
However, as we have no knowledge of the camera intrinsics, we employ two layers of 3D convolutions (without AdaIN) to morph the 3D representation into a perspective frustum (see Figure~\ref{fig:RenderGANArchitecture}) before their projection to 2D features.

The projection unit is composed of a reshaping layer that concatenates the channel dimension with the depth dimension, thus reducing the tensor dimension from 4D ($W \!\times\! H \!\times\! D \!\times\! C$) to 3D  ($W \!\times\! H \!\times\! (D \!\cdot\! C))$, and an MLP with a non-linear activation function (leakyReLU \cite{lRelu} in our experiments) to learn occlusion.
\subsection{Loss functions}
\label{sec:loss}
\paragraph{Identity regulariser}
To generate images at higher resolution (128$\times$128\,pixels), we find it beneficial to add an identity regulariser $L_\text{identity}$ that ensures a vector reconstructed from a generated image matches the latent vector $\mathbf{z}$ used in the generator G. 
We find that this encourages HoloGAN to only use $\mathbf{z}$ for the identity to maintain the object's identity when poses are varied, helping the model learn the full variation of poses in the dataset.
We introduce an encoder network $\text{F}$ that shares the majority of the convolution layers of the discriminator, but uses an additional fully-connected layer to predict the reconstructed latent vector.
The identity loss is:
\begin{align}
L_\text{identity}(\text{G}) &= \mathbb{E}_\mathbf{z} \norm{\mathbf{z} - \text{F}(\text{G}(\mathbf{z}))}^2 \text{.}
\end{align}

\paragraph{Style discriminator}
Our generator is designed to match the ``style'' of the training images at different levels, which effectively controls image attributes at different scales.
Therefore, in addition to the image discriminator that classifies images as real or fake, we propose multi-scale \emph{style discriminators} that perform the same task but at the feature level. 
In particular, the style discriminator tries to classify the mean $\mu(\mathbf{\Phi}_l)$ and standard deviation $\sigma(\mathbf{\Phi}_l)$, which describe the image ``style'' \cite{huang2017adain}.
Empirically, the multi-scale style discriminator helps prevent mode collapse and enables longer training.
Given a style discriminator $\text{D}_l(\mathbf{x}) \!=\! \widetilde{\text{D}}_l(\mu(\mathbf{\Phi}_l(\mathbf{x})), \; \sigma(\mathbf{\Phi}_l(\mathbf{x})))$ for layer $l$,
the style loss is defined as:
\begin{align}
L_\text{style}^l(\text{G}) &= \mathbb{E}_\mathbf{z} [-\log \; {\text{D}_l(\text{G}(\mathbf{z}))}] \text{.}
\end{align}
The total loss can be written as:
\begin{align}
L_\text{total}(\text{G}) &= L_\text{GAN}(\text{G})
+ \lambda_\text{i} \!\cdot\! L_\text{identity}(\text{G})
+ \lambda_\text{s} \!\cdot\! \sum_{l} L_\text{style}^l(\text{G})
\text{.}
\end{align}
We use $\lambda_\text{i} \!=\! \lambda_\text{s} \!=\! 1.0$ for all experiments. We use the GAN loss from DC-GAN \cite{Radford2016} for $L_{\text{GAN}}$.

\section{Experiment settings}

\paragraph{Data}
We train HoloGAN using a variety of datasets:
Basel Face \cite{Paysan2009},
CelebA \cite{liu2015faceattributes},
Cats \cite{zhang2008},
Chairs \cite{Chang2015},
Cars \cite{Yangg2015},
and LSUN bedroom \cite{yu15lsun}.
We train HoloGAN on resolutions of 64$\times$64\,pixels for Cats and Chairs, and 128$\times$128\,pixels for Basel Face, CelebA, Cars and LSUN bedroom. More details on the datasets and network architecture can be found in the supplemental document.

Note that only the Chairs dataset contains multiple views of the same object; all other datasets only contain unique single views.
For this dataset, because of the limited number of ShapeNet \cite{Chang2015} 3D chair models (6778 shapes), we render images from 60 randomly sampled views for each chair.
During training, we ensure that each batch contains completely different types of chairs to prevent the network from using set supervision, i.e., looking at the same chair from different viewpoints in the same batch, to cheat.

\paragraph{Implementation details}
We use adaptive instance normalization \cite{huang2017adain} for the generator, and a combination of instance normalization \cite{UlyanovVL17} and spectral normalization \cite{miyato2018spectral} for the discriminator. See our supplemental document for details.
 
We train HoloGAN from scratch using the Adam solver \cite{adam}.
To generate images during training, we sample $\mathbf{z} \!\sim\! \mathcal{U}(-1,1)$, and also sample random poses from a uniform distribution (more details on pose sampling can be found in the supplemental document).
We use $\abs{\mathbf{z}} \!=\! 128$ for all datasets, except for Cars at 128$\times$128, where we use $\abs{\mathbf{z}} \!=\! 200$.
Our code is available at \href{https://github.com/thunguyenphuoc/HoloGAN}{https://github.com/thunguyenphuoc/HoloGAN}.

\section{Results}
\label{sec:results}

We first show qualitative results of HoloGAN on datasets with increasing complexity
(Section \ref{sec:qualeval}).
Secondly, we provide quantitative evidence that shows HoloGAN can generate images with comparable or higher visual fidelity than other 2D-based GAN models (Section \ref{sec:quanteval}).
We also show the effectiveness of using our learnt 3D representation compared to explicit 3D geometries (binary voxel grids) for image generation (Section \ref{VON}).
We then show how HoloGAN learns to disentangle shape and appearance (Section \ref{sec:latentspace}).
Finally, we perform an ablation study to demonstrate the effectiveness of our network design and training approach (Section \ref{sec:ablation}).

\subsection{Qualitative evaluation}
\label{sec:qualeval}

\begin{figure}
	\centering
	\includegraphics[width=7.5cm]{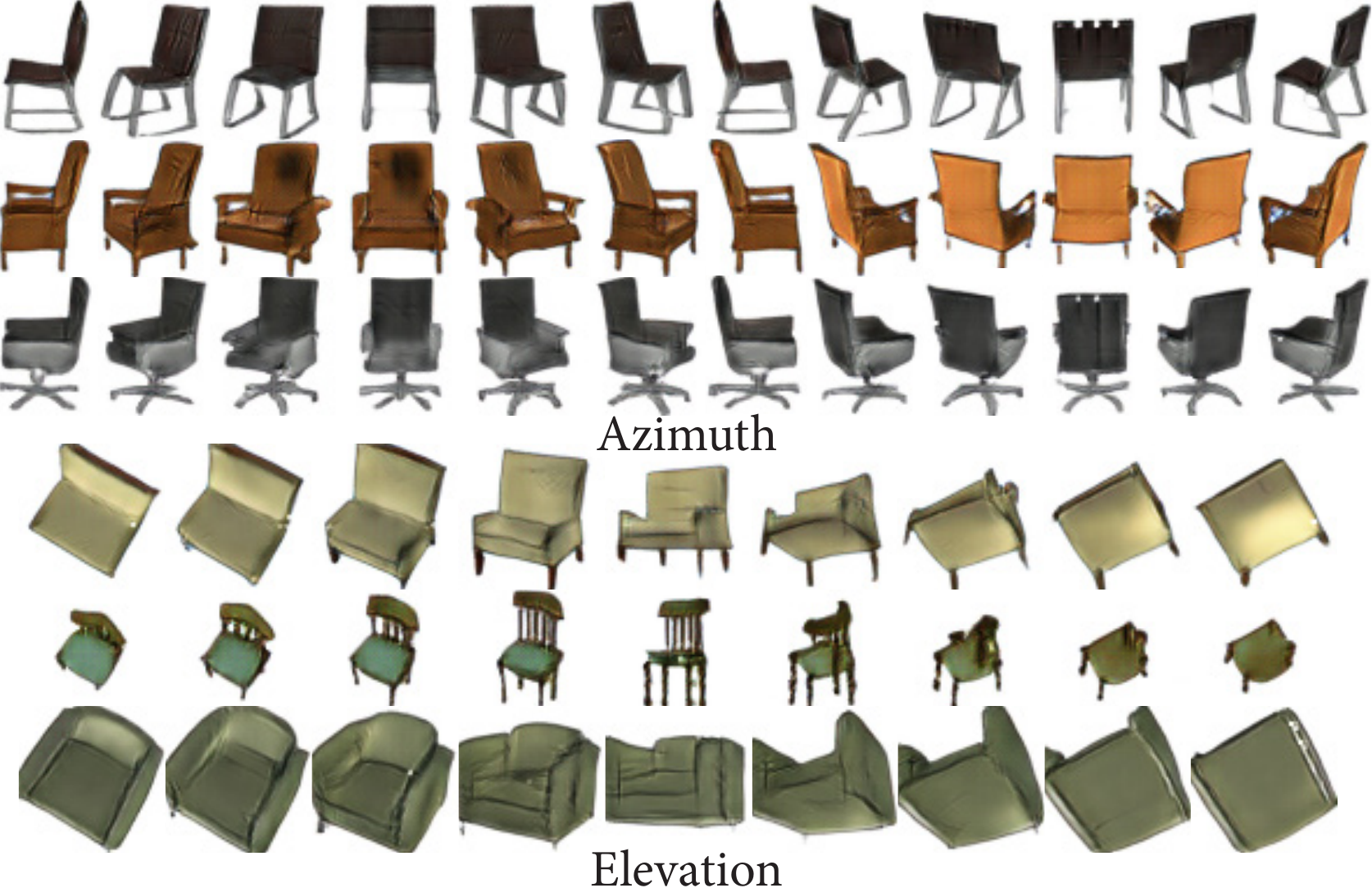}
	\caption{\label{fig:chairRotation}
		For the Chairs dataset with high intra-class variation, HoloGAN can still disentangle pose (360° azimuth, 160° elevation) and identity.
	}
\end{figure}

\begin{figure*}[!h]
	\centering
	\includegraphics[width=15.9 cm]{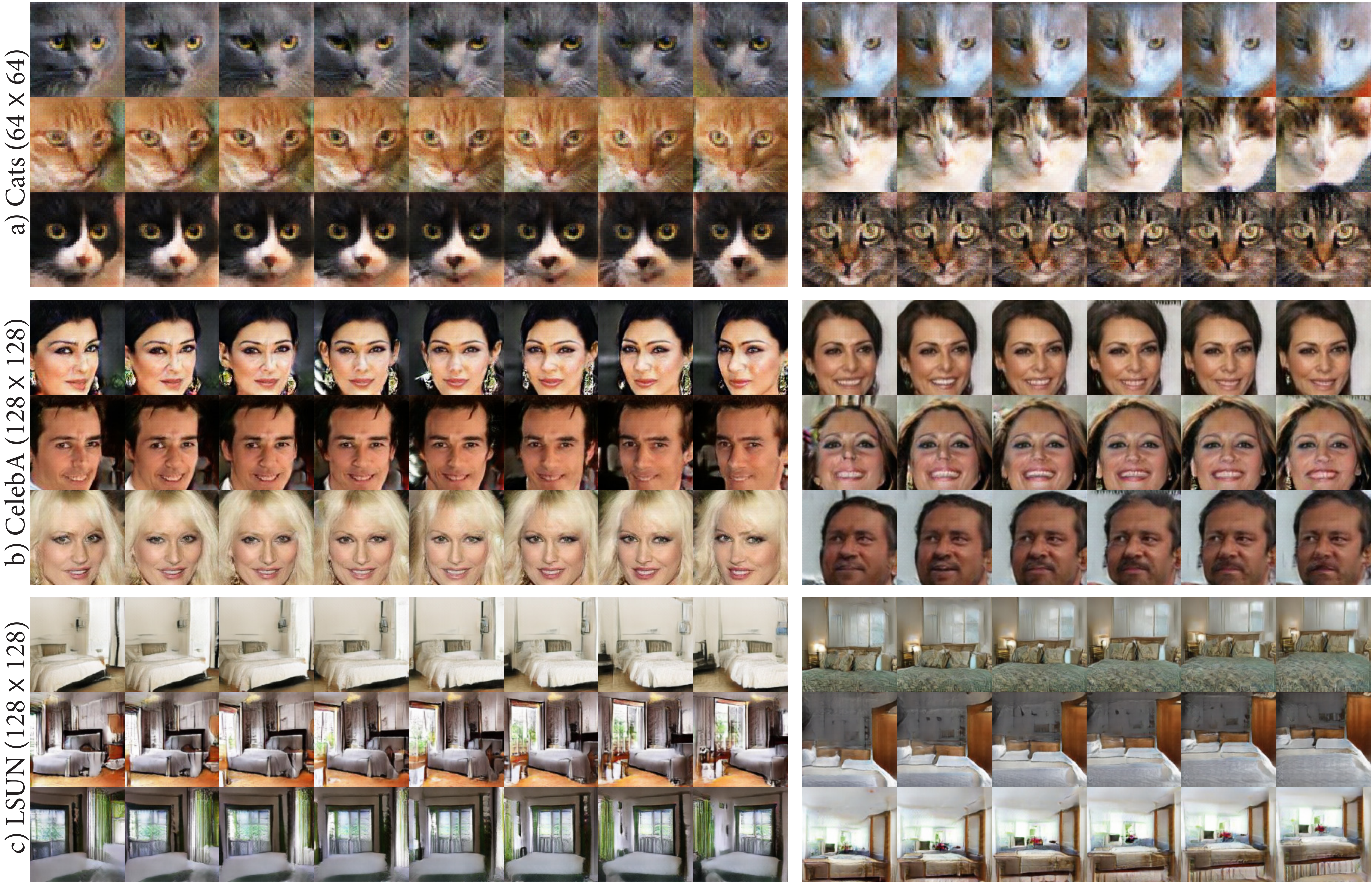}
	\caption{\label{fig:rotateFace}
		HoloGAN supports changes in both azimuth (range: 100°) and elevation (range: 35°). However, the available range depends on the dataset. For CelebA, for example, few photos in the dataset were taken from above or below.}
\end{figure*}

\begin{figure*}[!h]
	\centering
	\includegraphics[width=14.0cm]{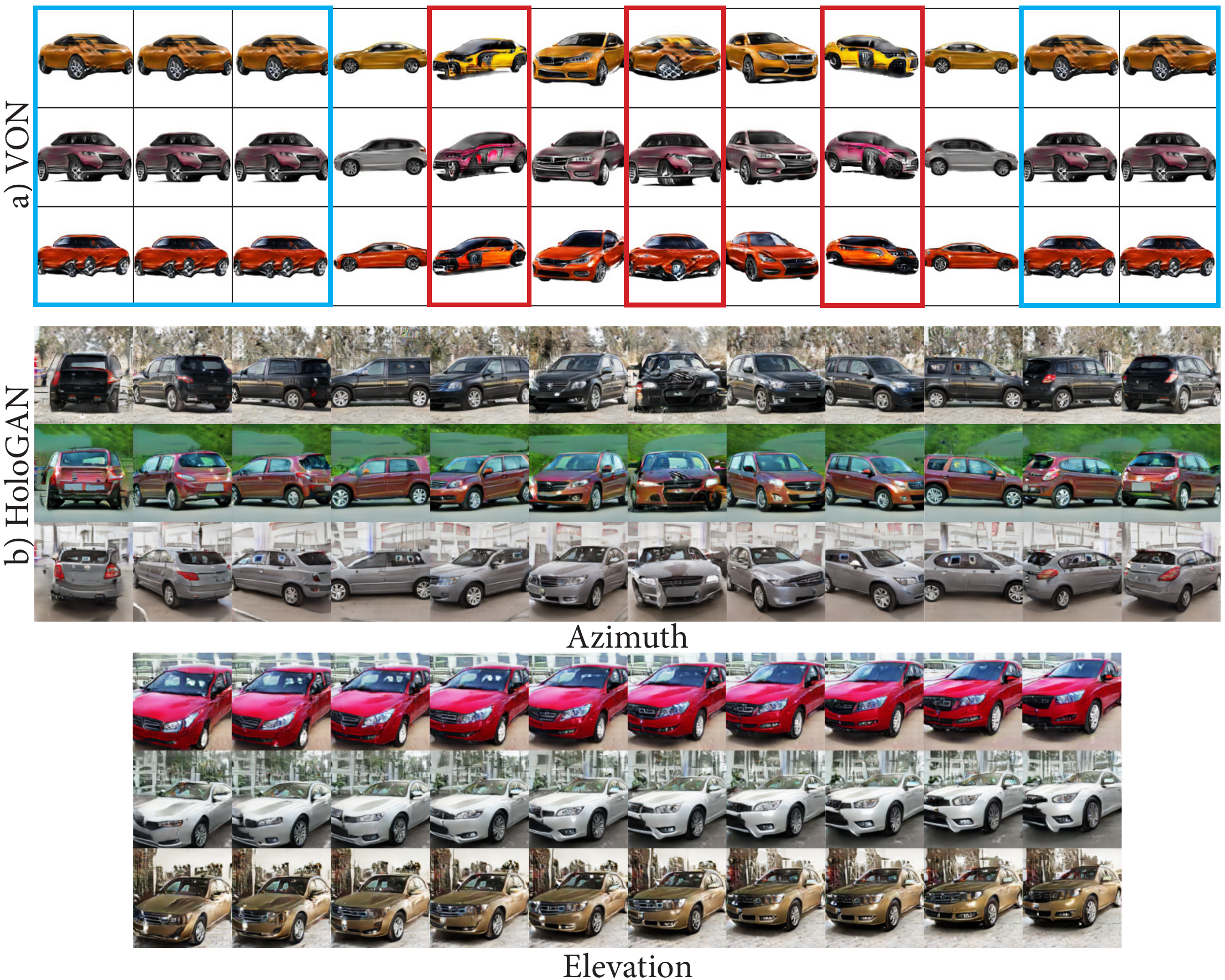}
	\caption{\label{fig:VON_compare}%
\textbf{a)} Car images generated by VON for an azimuth range of 360°.
Despite using 3D shapes and silhouette masks during training, VON can only generate images with a simple white background, and struggles for certain frontal views (highlighted in red) and rear views (highlighted in blue).
\textbf{b)} HoloGAN generates car images with complex backgrounds and varying azimuth (range: 360°) and elevation (range: 35°) from unlabelled images.}
\end{figure*}

Figures \ref{fig:moneyShot}, \ref{fig:chairRotation}, \ref{fig:rotateFace} and \ref{fig:VON_compare}b show that HoloGAN can smoothly vary the pose along azimuth and elevation while keeping the same identities for multiple different datasets.  
Note that the LSUN dataset contains a variety of complex layouts of multiple objects.
This makes it a very challenging dataset for learning to disentangle pose from object identity.

In the supplemental document, we show results for the Basel Face dataset. We also perform linear interpolation with the noise vectors while keeping the poses the same, and show that HoloGAN can smoothly interpolate the identities between two samples. This demonstrates that HoloGAN correctly learns an explicit deep 3D representation that disentangles pose from identity, despite not seeing any pose labels or 3D shapes during training.

\begin{figure}
\centering
\includegraphics[width=6.5cm]{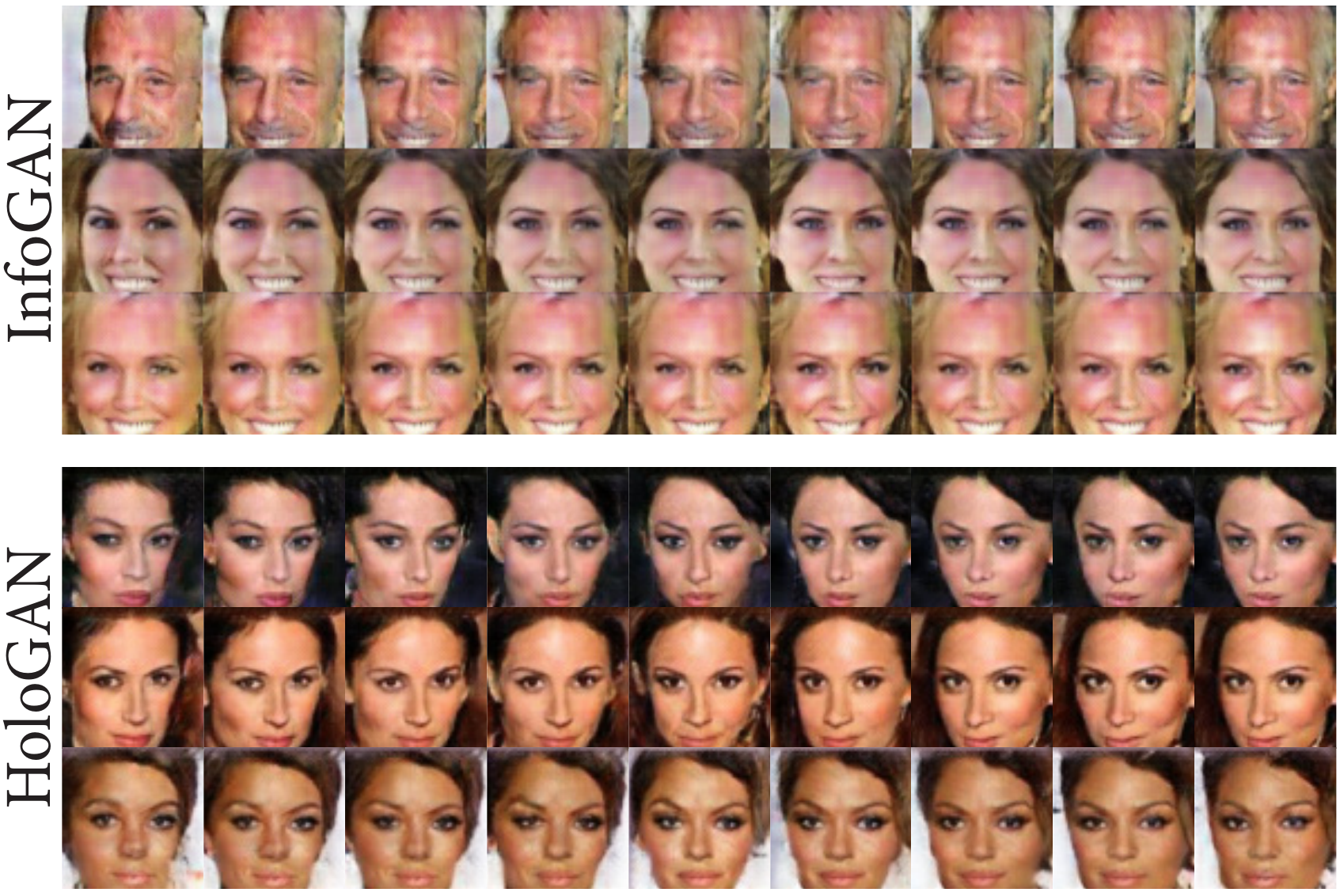}
\caption{\label{fig:compare_INFOGAN}%
	We compare HoloGAN to InfoGAN (images adapted from \citet{Chen2016}) on CelebA (64$\times$64) in the task of separating identity and azimuth.
	Note that we cannot control what can be learnt by InfoGAN. }
\end{figure}

\paragraph{Comparison to InfoGAN \cite{Chen2016}}
We compare our approach to InfoGAN on the task of learning to disentangle identity from pose on the CelebA dataset \cite{liu2015faceattributes} at a resolution of 64$\times$64 pixels.
Due to the lack of publicly available code and hyper-parameters for this dataset\footnote{The official code repository at \href{https://github.com/openai/InfoGAN}{https://github.com/openai/InfoGAN} only works with the MNIST dataset.}, we use the CelebA figure from the published paper.
We also tried the official InfoGAN implementation with the Cars dataset, but were unable to train the model successfully as InfoGAN appears to be highly sensitive to the choice of prior distributions and the number of latent variables to recover.

Figure \ref{fig:compare_INFOGAN} shows that HoloGAN successfully recovers and provides much better control over the azimuth while still maintaining the identity of objects in the generated images.
HoloGAN can also recover elevation (Figure \ref{fig:rotateFace}b, right) despite the limited variation in elevation in the CelebA dataset, while InfoGAN cannot.
Most importantly, there is no guarantee that InfoGAN always recovers factors that control object pose, while HoloGAN explicitly controls this via rigid-body transformation.

\subsection{Quantitative results}
\label{sec:quanteval}
To evaluate the visual fidelity of generated images, we use the Kernel Inception Distance (KID) by \citet{binkowski2018demystifying}\footnote{\href{https://github.com/mbinkowski/MMD-GAN}{https://github.com/mbinkowski/MMD-GAN}}.
KID computes the squared maximum mean discrepancy between feature representations (computed from the Inception model \cite{Inception}) of the real and generated images.
In contrast to FID \cite{NIPS2017_7240}, KID has an unbiased estimator.
The lower the KID score, the better the visual quality of generated images.
We compare HoloGAN with other recent GAN models: DCGAN \cite{Radford2016}, LSGAN \cite{Mao2017}, and WGAN-GP \cite{NIPS2017_7159}, on 3 datasets in Table \ref{table:KID}.
Note that KID does not take into account feature disentanglement, which is one of the main contributions of HoloGAN.

We use a publicly available implementation\footnote{\href{https://github.com/LynnHo/DCGAN-LSGAN-WGAN-WGAN-GP-Tensorflow}{https://github.com/LynnHo/DCGAN-LSGAN-WGAN-WGAN-GP-Tensorflow}} and use the same hyper-parameters (that were tuned for CelebA) provided with this implementation for all three datasets.
Similarly, for HoloGAN, we use the same network architecture and hyper-parameters\footnote{Except for ranges for sampling the azimuth: 100° for CelebA since face images are only taken from frontal views, and 360° for Chairs and Cars.} for all three datasets.
We sample 20,000 images from each model to calculate the KID scores shown below.

\begin{table}[h!]

\resizebox{\linewidth}{!}{%
\renewcommand*{\arraystretch}{1.2}%
\newcommand{\size}{{\footnotesize 64$\times$64}}%
\begin{tabular}{lccc}
	\toprule
	\textbf{Method}              &  \textbf{CelebA} \size   &  \textbf{Chairs} \size   &         \textbf{Cars} \size         \\ \midrule
	DCGAN \cite{Radford2016}     &     1.81 $\pm$ 0.09      &     6.36 $\pm$ 0.16      &     \phantom{0}4.78 $\pm$ 0.11      \\
	LSGAN \cite{Mao2017}         &     1.77 $\pm$ 0.06      &     6.72 $\pm$ 0.19      &     \phantom{0}4.99 $\pm$ 0.13      \\
	WGAN-GP \cite{NIPS2017_7159} & \textbf{1.63 $\pm$ 0.09} &     9.43 $\pm$ 0.24      &          15.57 $\pm$ 0.29           \\
	HoloGAN (ours)               &     2.87 $\pm$ 0.09      & \textbf{1.54 $\pm$ 0.07} & \phantom{0}\textbf{2.16 $\pm$ 0.09} \\ \bottomrule
\end{tabular}}\vspace{.5em}
\caption{\label{table:KID}%
	KID \cite{NIPS2017_7240} between real images and images generated by HoloGAN and other 2D-based GANs (lower is better).
	We report KID mean$\times$100 $\pm$ std.$\times$100.
	The table shows that HoloGAN can achieve competitive or higher KID score with other methods, while providing explicit control of objects in the generated images (not measured by KID).}
\end{table}

Table \ref{table:KID} shows that HoloGAN can generate images with competitive (for CelebA) or even better KID scores on more challenging datasets: Chairs, which has high intra-class variability, and Cars, which has complex backgrounds and lighting conditions. This also shows that the HoloGAN architecture is more robust and can consistently produce images with high visual fidelity across different datasets with the same set of hyper-parameters (except for azimuth ranges).
We include visual samples for these models in the supplemental document.
More importantly, HoloGAN learns a disentangled representation that allows manipulation of the generated images.
This is a great advantage compared to methods such as $\beta$-VAE \cite{higgins2017}, which has to compromise between the image quality and the level of disentanglement of the learnt features.

\subsection{Deep 3D representation vs. 3D geometry}
\label{VON}
Here we compare our method to the state-of-the-art visual object networks (VON) \cite{NIPS2018_7297} on the task of generating car images. We use the trained model and code provided by the authors. 
Although VON also takes a disentangling approach to generating images, it relies on 3D shapes and silhouette masks during training, while HoloGAN does not.
Figure \ref{fig:VON_compare}b shows that our approach can generate images of cars with complex backgrounds, realistic shadows and competitive visual fidelity.
Note that to generate images at 128$\times$128,  VON uses the full binary voxel geometry at 128$\times$128$\times$128$\times$1 resolution, while HoloGAN uses a deep voxel representation of up to 16$\times$16$\times$16$\times$64 resolution, which is more spatially compact and expressive since HoloGAN also generates complex backgrounds and shadows.
As highlighted in Figure \ref{fig:VON_compare}a, VON also tends to change the identity of the car such as changing colours or shapes at certain views (highlighted), while HoloGAN maintains the identity of the car in all views.
Moreover, HoloGAN can generate car images in full 360° views (Figure~\ref{fig:VON_compare}), while VON struggles to generate images from the back views.

Traditional voxel grids can be very memory intensive.
HoloGAN hints at the great potential of using explicit deep voxel representations for image generation, as opposed to using the full 3D geometry in the traditional rendering pipeline.
For example, in Figure~\ref{fig:rotateFace}c, we generate images of the entire bedroom scene using a 3D representation of only 16$\times$16$\times$16$\times$64 resolution.

\subsection{Disentangling shape and appearance}
\label{sec:latentspace}
\begin{figure}
	\centering
	\includegraphics[width=0.95\linewidth]{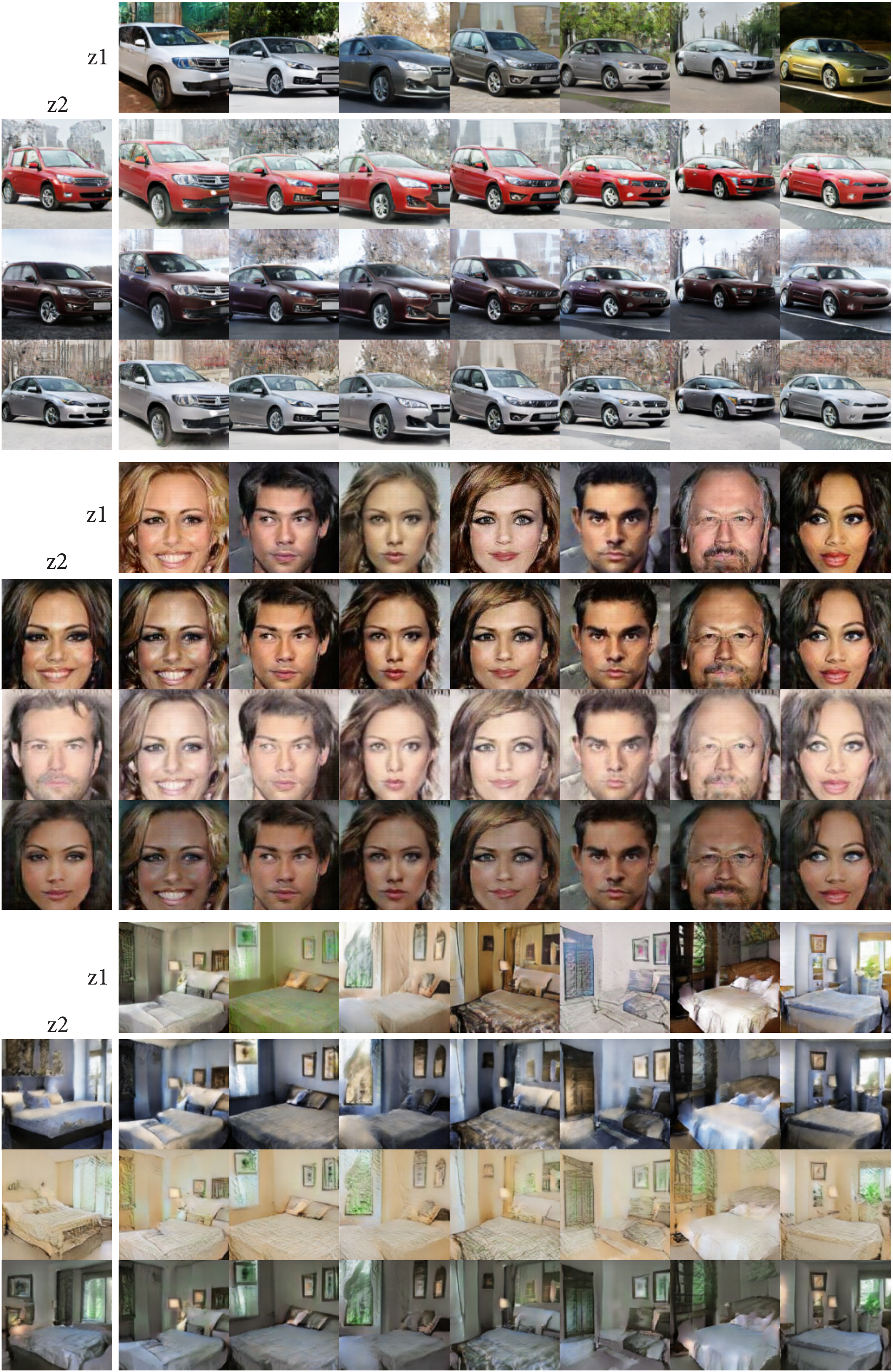}
	\caption{\label{fig:style_mix_1}%
		Combinations of different latent vectors $\mathbf{z}_1$ (for 3D features) and $\mathbf{z}_2$ (for 2D features).
		While $\mathbf{z}_1$ influences objects' shapes, $\mathbf{z}_2$ determines appearance (texture and lighting). Best viewed in colour.}
\end{figure}

Here we show that in addition to pose, HoloGAN also learns to further divide identity into shape and appearance.
We sample two latent codes, $\mathbf{z}_1$ and $\mathbf{z}_2$, and feed them through HoloGAN. While $\mathbf{z}_1$ controls the 3D features (before perspective morphing and projection), $\mathbf{z}_2$ controls the 2D features (after projection).
Figure \ref{fig:style_mix_1} shows the generated images with the same pose, same $\mathbf{z}_1$, but with a different $\mathbf{z}_2$ at each row.
As can be seen, while the 3D features control objects' shapes, the 2D features control appearance (texture and lighting). 
This shows that by using 3D convolutions to learn 3D representations and 2D convolutions to learn shading, HoloGAN learns to separate shape from appearance directly from unlabelled images, allowing separate manipulation of these factors.
In the supplemental document, we provide further results, in which we use different latent codes for 3D features at different resolutions, and show the separation between features that control the overall shapes and more fine-grained details such as gender or makeup.

\subsection{Ablation studies}
\label{sec:ablation}
We now conduct a series of studies to demonstrate the effectiveness of our network design and training approach.

\begin{figure}
	\centering
	\includegraphics[width=\linewidth]{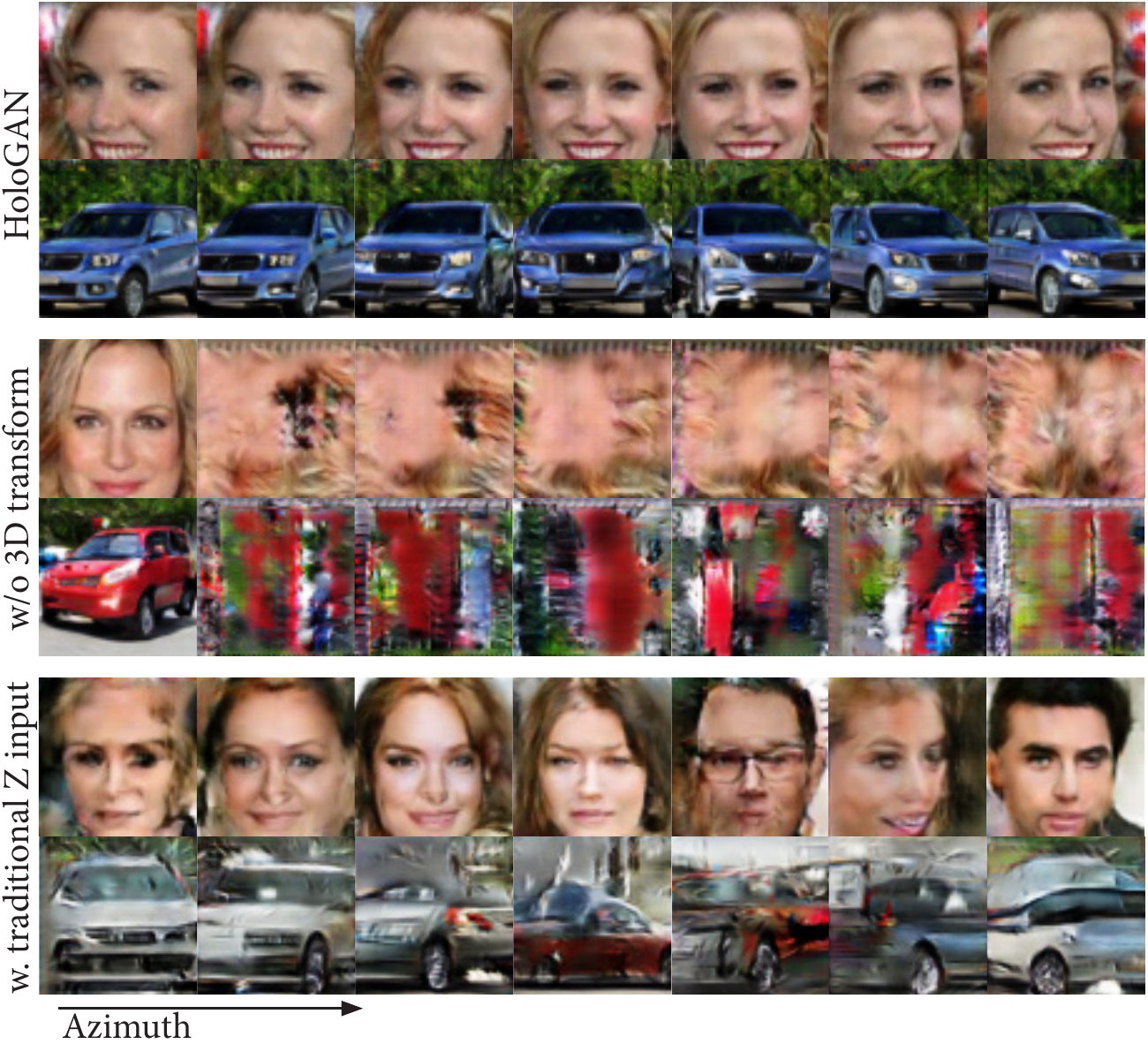}
	\caption{\label{fig:ablation}%
		Ablation study showing images with changing azimuths (from left to right).
		\textbf{Top:} Our approach.
		\textbf{Middle:} Our approach without using random 3D transformations during training fails to rotate objects.
		\textbf{Bottom:} Our approach with a traditional input layer mapped from $\mathbf{z}$ instead of a learnt constant tensor fails to disentangle object pose and identity.}
\end{figure}

\paragraph{Training without random 3D transformations}
Randomly rotating the 3D features during training is crucial for HoloGAN, as it encourages the generator to learn a disentangled representation between pose and identity.
In Figure \ref{fig:ablation}, we show results with and without 3D transformation during training.
For the model trained without 3D transformation, we generate images of rotated objects by manually rotating the learnt 3D features after the model is trained.
As can be seen, this model can still generate images with good visual fidelity, but when the pose is changed, it completely fails to generate meaningful images, while HoloGAN can easily generate images of the same objects in different poses.
We believe that the random transformation during training forces the generator to learn features that meaningfully undergo geometric transformations, while still being able to generate images that can fool the discriminator.
As a result, our training strategy encourages HoloGAN to learn a disentangled representation of identity and pose.
\paragraph{Training with traditional $\mathbf{z}$ input}
By starting from a learnt constant tensor and using the noise vector $\mathbf{z}$ as a ``style'' controller at different levels, HoloGAN can better disentangle pose from identity.
Here we perform another experiment, in which we feed $\mathbf{z}$ to the first layer of the generator network, like other GAN models.
Figure \ref{fig:ablation} shows that the model trained with this traditional input is confused between pose and identity.
As a result, the model also changes the object's identity when it is being rotated, while HoloGAN can smoothly vary the pose along the azimuth and keep the identity unchanged.

An additional ablation study showing the effectiveness of the identity regulariser is included in the supplemental document.

\section{Discussion and conclusion}

While HoloGAN can successfully learn to separate pose from identity, its performance depends on the variety and distribution of poses included in the training dataset.
For example, for the CelebA and Cats dataset, the model cannot recover elevation as well as azimuth (see Figure \ref{fig:rotateFace}a,b), since most face images are taken at eye level and thus contain limited variation in elevation. 
Using the wrong pose distribution might also lead to angles being mapped incorrectly.
Currently, during training, we sample random poses from a uniform distribution.
Future work therefore can explore learning the distribution of poses from the training data in an unsupervised manner to account for uneven pose distributions.
Other directions to explore include further disentanglement of objects' appearances, such as texture and illumination.
Finally, it will be interesting to combine HoloGAN with training techniques such as progressive GANs \cite{karras2018} to generate higher-resolution images.

In this work, we presented HoloGAN, a generative image model that learns 3D representation from natural images in an unsupervised manner by adopting strong inductive biases about the 3D world. HoloGAN can be trained end-to-end with only unlabelled 2D images, and learns to disentangle challenging factors such as 3D pose, shape and appearance. This disentanglement provides control over these factors, while being able to generate images with similar or higher visual quality than 2D-based GANs. 
Our experiments show that HoloGAN successfully learns meaningful 3D representations across multiple datasets with varying complexity.
We are therefore convinced that explicit deep 3D representations are a crucial step forward for both the interpretability and controllability of GAN models, compared to existing explicit (meshes, voxels) or implicit \cite{Eslami1204, rajeswar2019pixscene} 3D representations.

\paragraph{Acknowledgements}
We received support from the European Union's Horizon 2020 research and innovation programme under the Marie Skłodowska-Curie grant agreement No. 665992,
the EPSRC Centre for Doctoral Training in Digital Entertainment (EP/L016540/1),
RCUK grant CAMERA (EP/M023281/1),
an EPSRC-UKRI Innovation Fellowship (EP/S001050/1),
and an NVIDIA Corporation GPU Grant.
We thank Lambda Labs for GPU cloud credits and a travel grant.

{\small
\bibliographystyle{ieeenat_fullname}
\bibliography{egbib}
}

\end{document}